\documentclass[lettersize,journal,nosections]{IEEEtran}
\usepackage{amsmath,amsfonts}
\usepackage[linesnumbered,ruled,vlined]{algorithm2e}
\usepackage{array}
\usepackage[caption=false,font=normalsize,labelfont=sf,textfont=sf]{subfig}
\usepackage{textcomp}
\usepackage{stfloats}
\usepackage{url}
\usepackage{verbatim}
\usepackage{graphicx}
\usepackage{cite}
\usepackage{enumitem}
\usepackage{booktabs}
\usepackage{cleveref}
\crefformat{table}{#2Table~\Roman{#1}#3}
\usepackage[inkscapelatex=false]{svg}
\usepackage{threeparttable}
\usepackage{multirow}

\usepackage{fancyhdr} 
\usepackage[switch]{lineno}
\usepackage{xcolor} 

\begin{document}


\title{Modeling Relational Logic Circuits for And-Inverter Graph Convolutional Network}

\author{Weihao Sun, Shikai Guo, Siwen Wang, Qian Ma, Hui Li 


\IEEEcompsocitemizethanks{
                


   
                            }
			
			\thanks{}
		}





\maketitle
\begin{abstract}

The automation of logic circuit design enhances chip performance, energy efficiency, and reliability, and is widely applied in the field of Electronic Design Automation (EDA).
And-Inverter Graphs (AIGs) efficiently represent, optimize, and verify the functional characteristics of digital circuits, enhancing the efficiency of EDA development.
Due to the complex structure and large scale of nodes in real-world AIGs, accurate modeling is challenging, leading to existing work lacking the ability to jointly model functional and structural characteristics, as well as insufficient dynamic information propagation capability.
To address the aforementioned challenges, we propose AIGer, with the aim to enhance the expression of AIGs and thereby improve the efficiency of EDA development.
Specifically, AIGer consists of two components: 1) Node logic feature initialization embedding component and 2) AIGs feature learning network component.
The node logic feature initialization embedding component projects logic nodes, such as \texttt{AND} and \texttt{NOT}, into independent semantic spaces, to enable effective node embedding for subsequent processing.
Building upon this, the AIGs feature learning network component employs a heterogeneous graph convolutional network, designing dynamic relationship weight matrices and differentiated information aggregation approaches to better represent the original structure and information of AIGs.
The combination of these two components enhances AIGer’s ability to jointly model functional and structural characteristics and improves its message passing capability, thereby strengthening its expressive power for AIGs and enhancing the development efficiency of logic circuits. 
Experimental results indicate that AIGer outperforms the current best models in the Signal Probability Prediction (SSP) task, improving MAE and MSE by 18.95\% and 44.44\%, respectively. 
In the Truth Table Distance Prediction (TTDP) task, AIGer achieves improvements of 33.57\% and 14.79\% in MAE and MSE, respectively, compared to the best-performing models\footnote{https://github.com/ichont/AIGer}.

\end{abstract}

\begin{IEEEkeywords}
Electronic Design Automation, And-Inverter Graphs, Logic Circuit Design
\end{IEEEkeywords}

\section{\textbf{Introduction}}
\label{intro}

In the field of Electronic Design Automation (EDA), the modeling and representation of Boolean networks and their simplified form, And-Inverter Graphs (AIGs), can effectively enhance the efficiency of EDA tools and optimize circuit performance~\cite{cai2024parallel,jiang2024ml}.
The core idea of And-Inverter Graphs (AIGs) is to represent Boolean functions using only AND and Inverter nodes, simplifying the graph structure and improving operational efficiency.
AIGs capture circuit functionality through the topological connections of logic gates, such as \texttt{AND} and \texttt{NOT}, with their ability to represent functionality directly influencing the efficiency of critical tasks, including logic synthesis and formal verification~\cite{brayton2010abc}. 
Traditional approaches for handling AIGs typically rely on structural hashing and functional propagation techniques. 
However, these approaches are limited by heuristic rules and computational complexity, rendering them insufficient for the analysis of ultra-large-scale circuits~\cite{wu2023gamora,subramanyan2013reverse}.

\begin{figure}[!t]
\centering
 \includegraphics[width=0.95\linewidth]{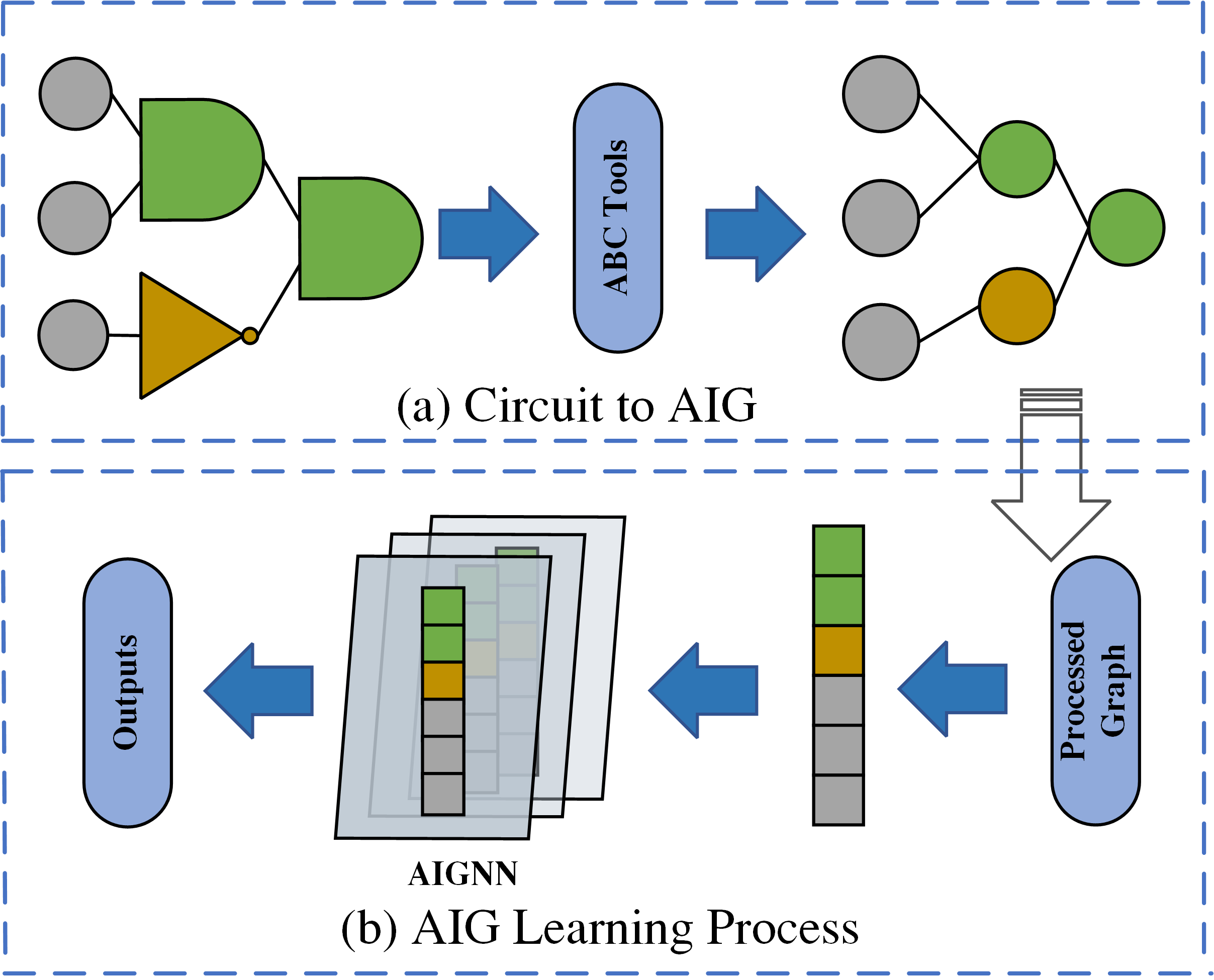}
\caption{The acquisition and learning process of AIG.}
\label{fig1}
\end{figure}

Deep learning has been widely applied in the field of EDA, establishing itself as a prominent research focus~\cite{huang2021machine,sanchez2023comprehensive,haaswijk2018deep}.
In recent years, AIG representation learning approaches based on Graph Neural Network (GNN), referred to as AIGNN, have achieved significant advancements~\cite{huang2021machine}.
To construct AIGs, researchers collect raw circuit data composed solely of \texttt{AND} gates, inverters, and input nodes, and transform it into corresponding AIG structures using ABC tools.
After conversion, node embeddings are initialized for the graph data, which is then fed into a pre-defined AIGNN model for graph learning. 
Through multiple rounds of message passing, the model produces task-specific outputs, thereby completing the learning and modeling process for AIGs.
~\figurename~\ref{fig1} illustrates the acquisition and learning process of AIG.
Previous AIGNN designs for processing AIGs have primarily targeted tasks with strong structural dependencies, such as congestion prediction~\cite{kirby2019congestionnet}, net length estimation~\cite{xie2021net2}, and power prediction~\cite{zhang2020grannite}.
FuncGNN~\cite{zhao2025funcgnn} adapts to structural heterogeneity by optimizing hybrid feature aggregation. It uses global normalization to balance gates and multi-layer integration to preserve global semantics, achieving efficient modeling of AIGs.
Given that AIG tasks are typically associated with the logical properties of nodes, several AIGNN approaches—such as DeepGate~\cite{li2022deepgate}, Polargate~\cite{liu2024polargate} and FuncGNN~\cite{zhao2025funcgnn} have enhanced embedding capabilities and demonstrated promising performance. 
These existing approaches employ supervised learning to encode the topological structures and node features of AIGs into a low-dimensional vector space, thereby establishing an efficient foundation for downstream inference tasks.

However, due to the complex structure and large scale of nodes in real-world AIGs, existing works fail to accurately represent and model them, facing the following two challenges:

\textbf{Challenge 1. Lacking the ability to jointly model functional and structural characteristics.}
The functional characteristics of AIGs generally refer to their Boolean logic semantics, specifically the logical properties of nodes; structural characteristics typically describe their hierarchical topological structure, including the number of nodes, edge connections, and graph depth.
In the initial node embedding phase, existing AIGNN approaches treat \texttt{AND} and \texttt{NOT} gates as homogeneous nodes, disregarding their functional characteristics, which leads to high embedding similarity and an inability to distinguish the functional differences between logic gates.
Prior work based on synchronous message passing has improved node-level functional representations and training efficiency; however, due to the lack of explicit modeling of circuit-level logic, these approaches fail to capture the topological depth of AIGs.
Although asynchronous message passing aligns well with circuit depth and better reflects the structural properties of AIGs, it often suffers from information compression issues due to excessive message-passing layers, leading to distorted functional and structural representations.

\textbf{Challenge 2. Lacking the sufficient capability for dynamic information propagation.}
Functional propagation in AIGs is path-sensitive, where the logical state of a node is determined by specific propagation paths. 
Due to the reliance of nodes in AIGs on both topological connections and functional propagation, existing approaches mainly focus on enhancing local structures with simple edges, which results in the inability to capture the path-dependent nature of functional propagation. 
This limitation leads to the indistinguishability of information during message passing, thereby affecting the overall performance of the model.
Additionally, during the aggregation phase, conventional operators like summation and averaging fail to account for gate-level distinctions, leading to a uniform aggregation that obscures the varying contributions of different logic paths.
Existing work also fail to distinguish between main paths and bypass signals in terms of weight allocation, leading to distorted functional representations under long-range dependencies. 
In practical scenarios, circuit structures often involve deep graph-based functional modeling, which necessitates increased AIGNN depth. 
However, deeper AIGNNs tend to suffer from over-compression of node feature and interaction information, resulting in information loss, while also incurring higher training time and computational cost.

To address the above challenges, we propose AIGer, which optimizes the modeling of AIGs, thereby improving the development efficiency of logic circuits.
Specifically, AIGer consists of the following two components: the node logic feature initialization embedding component and the AIGs feature learning network component.
In the node initialization embedding component, different nodes in the AIGs are represented using distinct embedding approaches, which aids in distinguishing the functional information of the nodes while preserving the semantic information of the initialized Boolean network. 
Building on this, the AIGs Feature Learning Network Component employs deep convolutional networks to construct dynamic weight matrices for each type of edge, while utilizing different aggregation approaches, thereby enhancing the overall modeling of AIGs' structural and functional characteristics.
Under the joint action of the two components, AIGer can address the potential issues at the current stage, effectively model and represent AIGs, thereby enhancing the development efficiency of logic circuits.

Experimental results on data sourced from four international standard circuit benchmark libraries show that AIGer outperforms the current best model by improving Signal Probability Prediction (SSP) tasks, with an 18.95\% improvement in MAE and a 44.44\% improvement in MSE.
In the Truth Table Distance Prediction (TTDP) task, AIGer achieves a 33.57\% improvement in MAE and a 14.79\% improvement in MSE compared to the current best model.
At the same time, the average training time of the model has also been optimized.

In summary, the main contributions of this work are:

\begin{itemize}
    \item We propose the AIGer model, with the Node Logic feature initialization embedding component to optimize the static representation of node functionality and structure, and the AIGs feature learning network component to enhance dynamic message passing, thereby better representing AIGs.

    \item The proposed AIGer model exhibits excellent capabilities in modeling and vectorizing representations of AIGs and Boolean networks, making it highly applicable to downstream tasks such as post-layout performance verification and power consumption prediction in the EDA field.

    \item The code and datasets for our work are publicly available~\cite{ourproject}, providing easy access for developers at any time.
\end{itemize}

\section{\textbf{Related work}}

\subsection{\textbf{And-inverter Graphs}}

Circuit representation learning is increasingly recognized as a key task in the EDA field, reflecting the broader trend of artificial intelligence's application in the electronics domain and its downstream tasks~\cite{shi2024deepgate3}.
AIGs are a simplified form of Boolean networks and are directed acyclic graphs (DAGs) used to represent the logical functions of digital circuits. Their core structure consists of \texttt{AND} gate nodes, inverters, and input nodes~\cite{biere2007aiger}.
Input nodes represent the original input signals of the circuit, and the final output is generated through a series of logical operations.

In practical applications, AIGs are widely used in circuit analysis, verification, and logic synthesis, among others~\cite{liu2024polargate}.
AIGs are well-suited for various downstream tasks in EDA, including functional reasoning and verification tasks, as well as design quality assessment tasks at the netlist stage, such as timing, power, and area estimation. In functional reasoning and verification, they help ensure the correctness of circuit functionality~\cite{fang2025survey}.
Researchers vectorize the representation of AIGs based on their characteristics for subsequent tasks. A key challenge in processing AIGs is simultaneously considering both the functional features, such as the types of logic gates in the circuit, and the structural features, such as the topological information.
Traditional approaches for processing AIGs include structural hashing and functional propagation~\cite{wu2023gamora,subramanyan2013reverse}; however, they often face challenges related to scalability and the efficient utilization of modern computational resources.
These challenges have prompted researchers to develop more advanced circuit representation learning models, particularly AIG representation models, thus driving the development of AIG representation learning.

\subsection{\textbf{Graph Learning in EDA}}
In recent years, Graph Neural Networks (GNNs) have become an effective approach for analyzing graph-structured data. The circuit design in EDA inherently possesses graph structural characteristics, where the connections between logic gates can be abstracted as a combination of nodes (logic gates) and edges (signal flows).
Research has shown that converting circuits into standardized graphs and representing them in this form can unify data formats across different design stages, thereby supporting performance prediction and optimization across toolchains~\cite{shrestha2024eda}.
The effectiveness of GNNs in EDA stems from their ability to model hierarchical dependencies within circuits, such as capturing long-range interactions in signal propagation paths~\cite{ren2022graph}.

In the field of EDA, researchers have proposed various customized and improved GNN models for different tasks and workflows.
Ustun et al.~\cite{ustun2020accurate} designed a GNN with generalization capabilities for learning circuit operation mapping patterns by distinguishing between predecessor and successor nodes in the graph.
The teams of Zhang~\cite{zhang2020grannite} and Guo~\cite{guo2022timing} respectively developed GNN models for power consumption inference and timing prediction, achieving task optimization through sequential updates of node representations.
Wu et al.~\cite{wu2023gamora} proposed a multi-task graph learning framework that supports circuit function reasoning.
Circuit-GNN maps the circuit to a graph network and performs backward optimization of high-frequency discrete circuit parameters to meet performance curve requirements~\cite{zhang2019circuit}.
Wang et al.~\cite{wang2022functionality} introduced graph contrastive learning to enhance generalization capability; however, the high computational cost impacts the model's scalability.
Although these models are task-specific and built upon the traditional message-passing paradigm~\cite{deng2024less}, they have demonstrated excellent results in their respective tasks.

\subsection{\textbf{AIG encoder Based on Deep Learning Techniques}}
Circuit encoding and deep learning models are capable of extracting key information from the structural and functional characteristics of AIGs and other logic circuits~\cite{cai2024parallel,jiang2024ml}. Deep learning techniques (DL) have shown good performance in extracting key information and achieving favorable results in AIG modeling~\cite{huang2021machine}.
AIG representation learning primarily uses the AIGNN encoding model, a GNN tailored for AIGs. Similar to traditional GNNs, the AIGNN model propagates node features along the original graph structure and transforms feature dimensions during the propagation process to ultimately generate low-dimensional node representations.
In recent years, many approaches have been developed that leverage graph learning techniques, encoding optimization, and other approaches, improving AIGs encoding and modeling. These advancements have significantly enhanced encoding accuracy and modeling efficiency for AIG structures, leading to breakthrough applications in key EDA tasks such as logic optimization and timing analysis~\cite{fang2025survey}.

The DeepGate family is one of the leaders in building circuit learning encoders~\cite{li2022deepgate,shi2023deepgate2,shi2024deepgate3,zheng2025deepgate4}. They use customized graph learning models to process AIGs. This series of models significantly enhances scalability by incorporating enhanced supervision mechanisms, optimized graph learning architectures, memory consumption management, and other improvement strategies.
DeepGate~\cite{li2022deepgate} proposed a GNN-based AIG encoding framework that simulates circuit logic behavior through attention mechanisms and bidirectional propagation.
DeepGate2~\cite{shi2023deepgate2} introduces the gate-level truth table Hamming distance as a supervisory signal and integrates functional semantics with structural topology embeddings.
DeepGate3~\cite{shi2024deepgate3} improves upon DeepGate2 by using a graph Transformer architecture, optimizing node embeddings and incorporating graph-level subgraph prediction tasks.
DeepGate4~\cite{zheng2025deepgate4} proposes a GAT-based sparse Transformer architecture for large-scale circuit design, combining hierarchical partitioning and structural encoding techniques to reduce computational complexity while ensuring the accuracy of circuit property learning.

\begin{figure*}[!t]
\centering
  \includegraphics[width=0.85\linewidth]{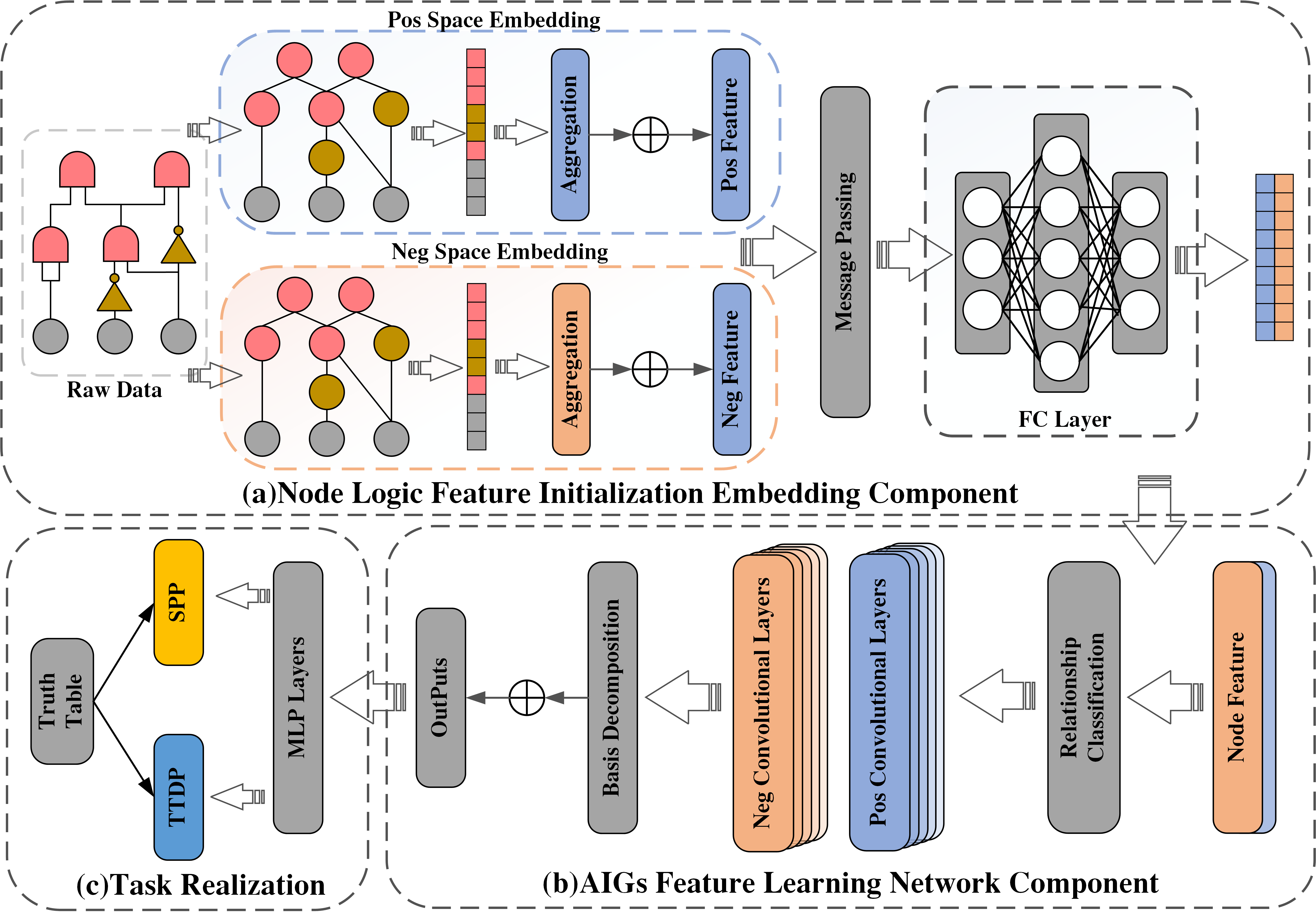}
\caption{The overall framework of AIGer}
\label{modelAIGer}
\end{figure*}

In addition to the DeepGate family, other works have also optimized custom GNN architectures, developed new message-passing mechanisms, and enhanced the scalability and performance of AIG encoding.
GAMORA~\cite{wu2023gamora} achieves high-precision and high-efficiency Boolean function analysis and arithmetic module identification in AIG modeling through a multitask learning framework combined with a symbolic reasoning mechanism.
HOGA~\cite{deng2024less} enhances the scalability and efficiency of AIG encoding by precomputing hop-wise features.
PolarGate~\cite{liu2024polargate} introduces a bipolar embedding space for modeling logical states in AIG, enhancing the expressive power of Boolean logic tasks.
The DeepSeq series~\cite{khan2024deepseq,khan2025deepseq2} extends AIG encoding to the domain of sequential logic circuits, using directed acyclic GNNs to process sequential netlists and improving the modeling accuracy of sequential logic.
The FGNN series~\cite{wang2022functionality,wang2024fgnn2} designs function-aware graph neural networks and introduces contrastive learning into the AIG encoding model. By aligning gate-level embeddings of positive and negative sample pairs through contrastive learning, it addresses the problems of arithmetic module recognition and functional similarity analysis.
Deepcell~\cite{shi2025deepcell} optimizes AIG encoding techniques by incorporating structural and functional information of AIG into the representation learning of PM netlists through a pretrained AIG encoder. It validates the bridging role of AIG in multimodal learning and empowers downstream tasks.
MGVGA~\cite{wu2025circuit} introduces large language models and addresses key challenges in circuit representation learning, such as preserving logical equivalence and abstract functional modeling, through a mask reconstruction mechanism with dual constraints on structure and functionality.
NetTAG~\cite{fang2025nettag} achieves a unified representation of AIG and other complex gate types through symbolic logical expressions and physical attribute annotations. It extends AIG encoding to general gate-level netlists for the first time, providing enhanced semantic understanding and physical design scalability for AIG encoding.
FuncGNN~\cite{zhao2025funcgnn} adapts to structural heterogeneity by optimizing hybrid feature aggregation. It uses global normalization to balance gates and multi-layer integration to preserve global semantics, achieving efficient modeling of AIGs.
These groundbreaking works continuously optimize the representational capacity of AIGNN and expand its application boundaries, providing an interpretable and scalable theoretical framework for AIG encoding and complex circuit topology modeling.

\section{\textbf{AIGer Model}}\label{Model}

In this section, we initially provide a comprehensive overview of AIGer. 
In Section A, we provide a brief description of AIGer. In the subsequent Section B and Section C, we introduce the key components of AIGer and explain how these components and structures handle AIGs during training and inference, as well as how they enhance the performance of AIGNN.

\subsection{\textbf{Overview}}\label{Overview}

To address the challenges of lacking the ability to jointly model functional and structural characteristics and lacking the sufficient capability for dynamic information propagation, we propose AIGer.
AIGer resolves these challenges by optimizing node embedding and information propagation, thereby improving the development efficiency of logic circuits.
The model architecture of AIGer is shown in~\figurename~\ref{modelAIGer}. 
The model consists mainly of two core components: the node logic feature initialization embedding component and the AIGs feature learning network component.
The Node Logic Feature Initialization Embedding Component represents different nodes in AIGs using distinct embedding approaches. 
By mapping them to separate node spaces based on their specific characteristics, it facilitates a better distinction of node features and Boolean semantics within the AIGNN. 
Building upon this, the AIGs Feature Learning Network Component constructs dynamic weight matrices for each edge type and applies distinct aggregation approaches to node features, further enhancing the model’s ability to capture both structural and functional information.
This enhances the model's ability to differentiate between the functionality and structure of AIGs, as well as improve message representation and passing. 
Under the joint action of the two components, AIGer can address the potential issues at the current stage, effectively model and represent AIGs, thereby enhancing the development efficiency of logic circuits.

\subsection{\textbf{Node Logic Feature Initialization Embedding Component}}
\label{embedding}

In this section, we address the node logic feature initialization embedding component, which effectively distinguishes and integrates the logical attributes of the nodes. 
This approach ensures a more accurate and efficient representation of AIGs, enhancing the overall modeling process, thus resolving Challenge 1.

In order to differentiate the functional characteristics of the nodes, i.e., the logical attributes of circuit nodes, during the initial embedding process of AIGs, AIGer performs a preliminary extraction of the node features of the AIGs and embed them into a separate semantic space to distinguish the independent features of the nodes. 
By distinguishing the positive and negative polarities of the logical states, the operational characteristics of the logic gates are explicitly encoded.
Considering that the nodes in AIGs include \texttt{AND} nodes, \texttt{NOT} nodes and input nodes, each node has distinct functions but possesses two logical states, 0 and 1.
AIGer first independently maps the nodes to two different semantic spaces corresponding to the two logical states. 
Then, the initialization embedding operation is performed using different logical operators within these separate semantic spaces.

Specifically, for the \texttt{AND} node, during the positive embedding representation, the initial feature of the node itself is weighted and summed with the features of neighboring nodes to obtain the positive embedding \( h^{\mathrm{Po(1)}} \). 
A linear transformation of the node features is performed using the feature matrix \( w^{\mathrm{Po}(1)} \), and the activation function \( \sigma \) is applied.
Positive embedding captures the logical characteristics of the \texttt{AND} gate, where the output is the intersection of the input signals. 
It reflects the logical function by aggregating the information from neighboring nodes. 
The formula is expressed as follows:

\begin{equation}\label{1}
h_i^{\mathrm{Po}\left(1\right)}=\sigma\left(w^{\mathrm{Po}\left(1\right)}\left[\sum_{j\in\mathcal{N}_i}\frac{h_j^{\left(0\right)}}{\left|\mathcal{N}_i\right|},h_i^{\left(0\right)}\right]\right)
\end{equation}
where \( h_i^{(0)} \) and \( h_j^{(0)} \) are the initial features of nodes \( i \) and \( j \), respectively. 
\( \mathcal{N}_i \) represents the set of neighbors of node \( i \), and \( w^{\mathrm{Po}(1)} \) is a learnable weight matrix in the positive embedding space. 
For the negative embedding representation of the \texttt{AND} node, the input feature is only the feature \( h_i^{(0)} \) of the node itself. 
A zero vector is concatenated with the feature \( h_i^{(0)} \) to form the input vector, which is then transformed by a learnable matrix.
Negative embedding does not introduce neighboring information and simulates the logical behavior of the \texttt{AND} gate in a negative state, such as the default output when input signals are missing. 
The specific formula is expressed as follows:

\begin{equation}\label{2}
h_i^{\mathrm{Ne}\left(1\right)}=\sigma\left(w^{\mathrm{Ne}\left(1\right)}\left[0,h_i^{\left(0\right)}\right]\right)
\end{equation}
where \( w^{\mathrm{Ne}(1)} \) is a learnable weight matrix in the negative embedding space, and \( h_i^{(0)} \) represents the feature of the input node itself, which is concatenated with a zero vector.

For the \texttt{NOT} node, the positive embedding process is similar to the negative embedding process of the \texttt{AND} node. 
The input consists only of the node's own feature \( h_i^{(0)} \), and after concatenation with the zero vector, a linear transformation and non-linear activation are applied. 
The positive embedding of the \texttt{NOT} gate does not depend on neighbor information, reflecting its logical inversion function. The equation is expressed as follows:

\begin{equation}\label{3}
h_i^{\mathrm{Po}\left(1\right)}=\sigma\left(w^{\mathrm{Po}\left(1\right)}\left[0,h_i^{\left(0\right)}\right]\right)
\end{equation}
where \( w^{\mathrm{Po}(1)} \) is a learnable weight matrix in the positive embedding space, and \( h_i^{(0)} \) represents the feature of the input node itself, which is concatenated with a zero vector.

For the negative embedding representation of the \texttt{NOT} node, similar to the positive embedding process of the \texttt{AND} node, the initial feature \( h_i^{(0)} \) of the node itself is weighted and summed with the initial features \( h_k^{(0)} \) of neighboring nodes, then concatenated. 
A linear transformation is applied using the weight matrix, followed by the activation function. 
The negative embedding simulates the inversion logic of the \texttt{NOT} gate through neighbor information. 
The equation is expressed as follows:

\begin{equation}\label{4}
h_i^{\mathrm{Ne}\left(1\right)}=\sigma\left(w^{\mathrm{Ne}\left(1\right)}\left[\sum_{k\in\mathcal{N}_i}\frac{h_k^{\left(0\right)}}{\left|\mathcal{N}_i\right|},h_i^{\left(0\right)}\right]\right)
\end{equation}
where \( h_i^{(0)} \) and \( h_k^{(0)} \) are the initial features of nodes \( i \) and \( k \), respectively. \( \mathcal{N}_i \) is the set of neighbors of node \( i \), and \( w^{\mathrm{Ne}(1)} \) is a learnable weight matrix in the positive embedding space.

For the initial embedding of input node \texttt{PI}, since there are no predecessor nodes, it is similar to the negative embedding of the \texttt{AND} node and the positive embedding of the \texttt{NOT} node. 
Both the positive and negative embeddings of \texttt{PI} only require its own initial feature and concatenation with a zero vector. 
The equation expression is the same as in Equations (2) and (3). As the input source of the logical network, the positive and negative embeddings of the \texttt{PI} node reflect only its own state and do not rely on neighbor information.
In the above embedding process, all nodes of the same type share the same parameter matrices \( w^{\mathrm{Po}(1)} \) and \( w^{\mathrm{Ne}(1)} \), ensuring the consistency of the embedding space. 
The activation function \( \tanh \) introduces non-linearity, This transformation allows the neural network to learn complex patterns and relationships.

The node embedding initialization component of AIGer explicitly encodes the functional characteristics of logic gates into the initial embeddings, laying the foundation for message aggregation and transmission in subsequent AIGs feature learning network component.

\subsection{\textbf{AIGs Feature Learning Network Component}}

In this section, we utilize the AIGs feature learning network component based on heterogeneous graph convolutional networks to address the issue of lacking sufficient capability for dynamic information propagation.
We optimized the network propagation and learning strategies, thus resolving Challenge 2.

\subsubsection{\textbf{AIGs Convolution Operation}}

AIGs typically involve deeper graph structures and possess heterogeneous characteristics. 
We adopt the weight learning pattern of R-GCN~\cite{schlichtkrull2018modeling}, which distinguishes node relationships by constructing different learnable weight matrices for different edges, addressing the challenge of heterogeneous characteristics in AIGs.

\begin{figure}[!t]
\centering
  \includegraphics[width=0.95\linewidth]{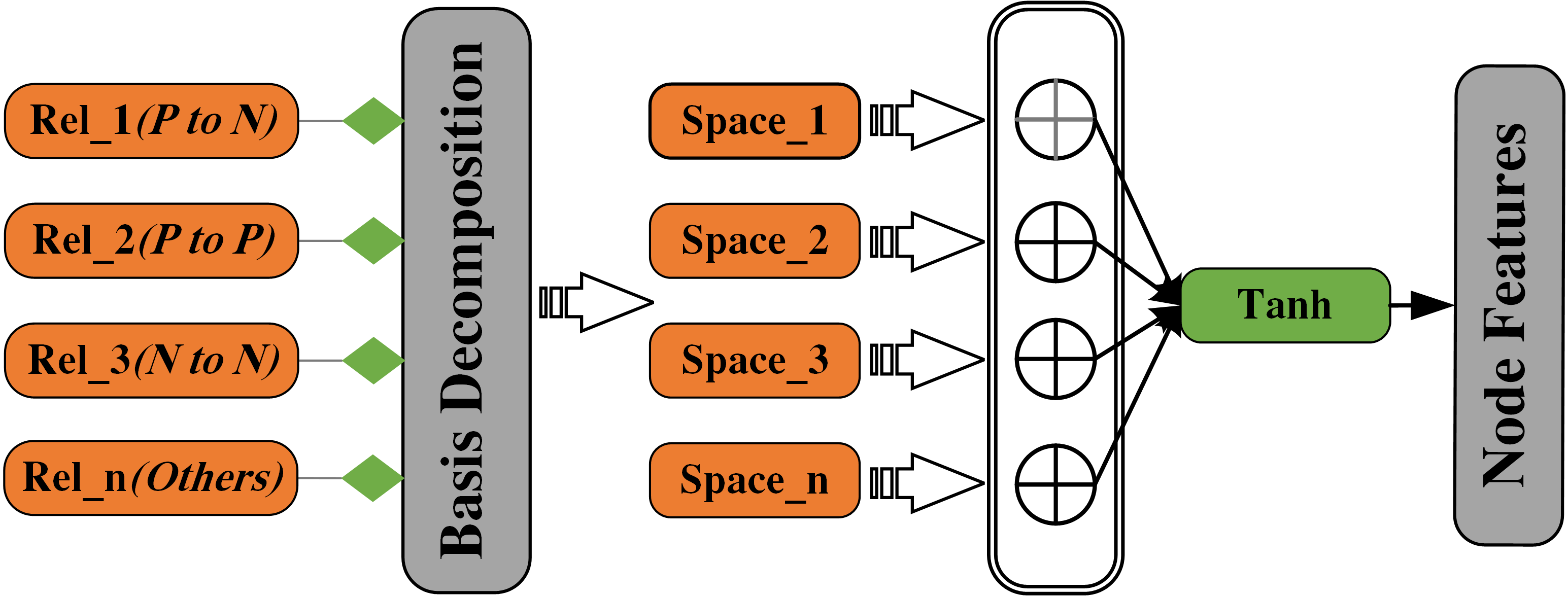}
\caption{The AIGs Feature Learning Network Component of AIGer}
\label{rgcn}
\end{figure}

In AIGNN, the node characteristics, Boolean semantics, and relationships between nodes, particularly structural features, are better learned. 
~\figurename~\ref{rgcn} shows the feature aggregation pattern of the convolution layer, which uses different aggregation operations for different edge types to effectively learn AIGs' structural functional characteristics and complete feature memory and propagation.
During the learning process, different edge types correspond to distinct relationships.

After the initialization embedding operation, the graph data has been embedded into different semantic spaces. 
During the forward propagation of the convolutional layer, the positive and negative embedding spaces adopt different convolutional learning modes. 
For the $l$-th layer, the positive embedding of node $n_i$ is updated as follows:

\begin{equation}\label{5}
\resizebox{0.44\textwidth}{!}{$
h_i^{Po\left(l+1\right)}=\sigma\left(\sum_{r\in\mathcal{R}}{\sum_{j\in N_i^r}\frac{1}{c_{i,r}}W_r^{Po\left(l\right)}h_j^{Po\left(l\right)}}+W_0^{Po\left(l\right)}h_i^{Po\left(l\right)}\right)
$}
\end{equation}
where \( N_i^r \) represents the set of neighboring nodes of node \( i \) under relation \( r \), where \( r \in \mathcal{R} \). \( c_{i,r} \) is a predefined normalization coefficient, which can also be learned. In our work, \( c_{i,r} \) is set to a fixed value equal to \( |N_i^r| \). 
The above equation aggregates the feature vectors of neighboring nodes through normalized summation, and the convolution operation depends on the direction and type of the edges. 
The update of the AIGer convolution layer neural network is performed in parallel, efficiently implemented via sparse matrix multiplication. 
Stacking multiple convolution layers allows the expression of graph embedding features and complex dependencies between nodes, while achieving richer and more precise message updates and transmission.

Similarly, using the same convolution operation, the negative embedding of node \( n_i \) is updated as:
\begin{equation}\label{6}
\resizebox{0.44\textwidth}{!}{$
h_i^{Ne\left(l+1\right)}=\sigma\left(\sum_{r\in\mathcal{R}}{\sum_{j\in N_i^r}\frac{1}{c_{i,r}}W_r^{Ne\left(l\right)}h_j^{Ne\left(l\right)}}+W_0^{Ne\left(l\right)}h_i^{Ne\left(l\right)}\right)
$}
\end{equation}

When applying graph data with a large number of nodes and edge types, especially complex graph data represented by AIGs, the number of parameters in the above equation increases dramatically. This leads to a large model size, low training efficiency, and makes it difficult to apply to practical problems. 
To reduce the parameter count in multi-relation scenarios, we employ a factorization strategy to optimize model training and inference efficiency, enabling adaptation to larger and more complex application scenarios under limited computational resources.
The equation is expressed as follows:

\begin{equation}\label{7}
W_r^{Po\left(l\right)}=\sum_{b=1}^{B}{a_{rb}^{Po\left(l\right)}V_b^{Po\left(l\right)}}
\end{equation}
\begin{equation}\label{8}
W_r^{Ne\left(l\right)}=\sum_{b=1}^{B}{a_{rb}^{Ne\left(l\right)}V_b^{Ne\left(l\right)}}
\end{equation}
where \( V_b^{\mathrm{Po}(l)} \) and \( V_b^{\mathrm{Ne}(l)} \) are shared matrices, which can be shared across the many relation types, reducing parameters while preserving the unique relationships of rare types. 
\( a_{rb}^{\mathrm{Po}(l)} \) and \( a_{rb}^{\mathrm{Ne}(l)} \) are relation-specific learnable coefficients, and \( B \) represents the parameter count of the base matrix. This strategy effectively addresses the issues of large parameter sizes and overfitting under large-scale AIG data, making AIGNN more applicable in practical work. 
Each relation type \( r \), including \texttt{AND}, \texttt{NOT}, and \texttt{PI} logical relations, corresponds to independent linear transformations and achieves cross-relation parameter interaction through shared base matrices while retaining relation-specific coefficients to ensure independent modeling of logical relations like \texttt{AND}, \texttt{NOT}, and \texttt{PI}.

\subsubsection{\textbf{Node Information Propagation Mechanism}}

To further optimize the network's message passing capability, we introduce a hierarchical aggregation operation to update feature information. 
During convolutional processing, linear information aggregation has already been implemented; 
however, this is insufficient to distinguish the logical characteristics of nodes. 
After the convolutional layers, we perform further aggregation of the node features to account for the functional differences brought by node types. Specifically, for the \texttt{AND} node, the message aggregation operation is as follows:

\begin{equation}\label{9}
\resizebox{0.44\textwidth}{!}{$
m_i^{AND}=\bigoplus_{r\in\{AND\}}\left(\frac{1}{\left|\mathcal{N}_{i}^{r}\right|}\sum_{j\in\mathcal{N}_{i}^{r}}{W_r^{Po}h_j^{Po}}\odot W_r^{Ne}h_j^{Ne}\right)
$}
\end{equation}
where \( \odot \) represents the Hadamard product, simulating the truth table characteristics of the Boolean AND operation, and \( \bigoplus \) denotes channel concatenation, retaining multi-relation interaction information. Correspondingly, for \texttt{NOT} nodes, the message aggregation operation formula is:

\begin{equation}
\resizebox{0.42\textwidth}{!}{$
m_i^{\text{NOT}} = \sum_{r \in \{ \text{NOT} \}} 
\left[ W_r^{\text{Ne}} \left( \max_{j \in \mathcal{N}_i^r} h_j^{\text{Po}} \right) 
- W_r^{\text{Po}} \left( \min_{j \in \mathcal{N}_i^r} h_j^{\text{Ne}} \right) \right]
$}
\end{equation}

In the above equation, the max operation captures the decisive features of logical negation, while the min operation suppresses negative information. 
After performing different aggregation operations for \texttt{AND} and \texttt{NOT} nodes, the node state update is carried out, integrating the message passing results with the node's own state. The final node state update equation for AIGer is:

\begin{equation}\label{11}
h_i^{Po\left(l+1\right)}=\sigma\left(W_\mathrm{r}^{Po\left(l\right)}h_i^{Po\left(l\right)}+m_i^{\mathrm{AND}}\right)
\end{equation}
\begin{equation}\label{12}
h_i^{Ne\left(l+1\right)}=\sigma\left(W_\mathrm{r}^{Ne\left(l\right)}h_i^{Ne\left(l\right)}+m_i^{\mathrm{NOT}}\right)
\end{equation}
where \( W_\mathrm{r}^{\mathrm{Po}(l)} \) and \( W_\mathrm{r}^{\mathrm{Ne}(l)} \) are relation-specific weights, associated with the base factorization strategy to ensure parameter efficiency. 
The hierarchical message passing mechanism designed for AIGer adapts to more complex logical gate combinations.

This design integrates relational graph convolution with Boolean logic priors, enabling AIGer to effectively represent AIGs, while enhancing the interpretability of functional representations through physics-guided aggregation operations.

\section{\textbf{Experimental Setup}}

In this section, we provide a comprehensive overview of the experimental setup for AIGer, including the evaluation strategies and metrics, the datasets and existing State-of-The-Art(SOTA) approaches used, as well as the training strategies and details.

\subsection{\textbf{Evaluation Metrics}}\label{pniggu}

To effectively test and evaluate the functional representation capability of AIGer, two tasks frequently used in previous works were selected: Signal Probability Prediction (SPP) and Truth Table Distance Prediction (TTDP). 
SPP is an important task in AIG representation learning and is widely used to assess model performance. 
TTDP, proposed based on DeepGate2~\cite{shi2023deepgate2}, is a novel fine-grained evaluation strategy used to assess the functional representation capability of GNNs.

For the Signal Probability Prediction (SPP) task, the core objective is to compute the mean absolute error between the true simulated logic output \( \widehat{y_v} \) of all nodes in the circuit and the predicted value \( y_v \) from the AIGNN model. 
Specifically, by traversing each node \( v \in V \) in the circuit network, we calculate the average error between the actual output and the predicted result of each node. The final evaluation metric formula is as follows:

\begin{equation}\label{11}
\mathrm{MAE.\ }{\mathrm{Error}}_{\mathrm{SPP}}=\frac{1}{N}\sum_{v\in V}\left|y_v-\widehat{y_v}\right|
\end{equation}
\begin{equation}\label{12}
\mathrm{MSE.\ }{\mathrm{Error}}_{\mathrm{SPP}}=\frac{1}{N}\sum_{v\in V}\left(y_v-\widehat{y_v}\right)^2
\end{equation}

In the Truth Table Distance Prediction (TTDP) task, the main focus is on measuring the functional similarity between nodes. If two nodes \( i \) and \( j \) have similar logical functions, their embedding vectors \( z_i \) and \( z_j \) should exhibit similarity. 
This relationship is modeled through the truth table vectors \( T_i \) and \( T_j \) of the nodes, with the specific mathematical relationship given as follows:
\begin{equation}\label{11}
\mathrm{distance}\left(z_i,z_j\right)\propto\mathrm{distance}\left(T_i,T_j\right)
\end{equation}

For the implementation details of the Truth Table Distance Prediction (TTDP) task, the specific process is as follows: 
First, by sampling a set of node pairs \( \mathcal{N} \), the Hamming distance \( D_{(i,j)}^T \) between the truth tables of each node pair \( (i,j) \) is calculated. 
It is defined as the ratio of the number of differing bits to the total length of the truth table, as shown in the following formula:

\begin{equation}\label{11}
D_{\left(i,j\right)}^T=\frac{\mathrm{HammingDistance} \left(T_i,T_j\right)}{\mathrm{|length}\left(T_i\right)|}
\end{equation}

The above formula reflects the relative degree of functional difference between nodes through the standardized Hamming distance. 
Based on the positive correlation between the embedding space and the truth table distance proposed in the previous formula, we further use cosine similarity to measure the similarity between the embedding vectors \( z_i \) and \( z_j \):

\begin{equation}\label{11}
D_{\left(i,j\right)}^Z=1-\mathrm{CosineSimilarity}\left(z_i,z_j\right).
\end{equation}

This design retains the symbolic-level semantic association through geometric space mapping. 
To eliminate the effect of dimensionality, after performing zero-norm normalization on the two types of distances \( D_{(i,j)}^T \) and \( D_{(i,j)}^Z \), we obtain \( D_{(i,j)}^{T^\prime} \) and \( D_{(i,j)}^{Z^\prime} \), respectively. 
The formulas for the mean absolute error and mean square error in the TTDP task are defined as follows:

\begin{equation}\label{11}
\mathrm{MAE.\ }{\mathrm{Error}}_{\mathrm{TTDP}}=\frac{1}{N}\sum_{\left(i,j\right)\in\mathcal{N}}\left|D_{\left(i,j\right)}^{T^\prime}-D_{\left(i,j\right)}^{Z^\prime}\right|
\end{equation}
\begin{equation}\label{12}
{\mathrm{MSE.\ Error}}_{\mathrm{TTDP}}=\frac{1}{N}\sum_{\left(i,j\right)\in\mathcal{N}}\left(D_{\left(i,j\right)}^{T^\prime}-D_{\left(i,j\right)}^{Z^\prime}\right)^2
\end{equation}

We select Mean Absolute Error (MAE), Mean Squared Error (MSE), and Average Running Time (Avg.RT) as core metrics. Both MAE and MSE are suitable for quantifying the single-node prediction bias in the SPP task and aggregating multi-node relational errors in the TTDP task. 
Parameter updates are achieved by directly minimizing prediction differences. 
Additionally, we monitor the model's training time to assess its practical value under limited computational resources. 
\( T_{\text{avg}} \) represents the average time per training epoch, calculated by the model's error convergence under the MSE metric, with the unit in seconds. Training time must be compared within the same device and experimental environment.

\subsection{\textbf{Datasets}}

In this paper, we utilize the dataset prepared by Liu et al. \cite{liu2024polargate}.
The original data is sourced from four international standard circuit benchmark libraries: ITC'99\cite{davidson1999characteristics}, IWLS'05\cite{albrecht2005iwls}, EPFL\cite{amaru2015epfl}, and OpenCore\cite{shujuji4}. 
After converting the circuits from each benchmark library into AIGs structured representations using a unified logic synthesis tool, a standardized dataset containing multi-level logic gate circuits is created. 
The dataset composition is shown in Table~\ref{dataset}. 
Its statistical features exhibit significant heterogeneity, with node sizes ranging from tens to thousands, and notable differences in logic depth, effectively capturing the complexity distribution characteristics in real-world digital integrated circuit designs.

\subsection{\textbf{SOTA Approaches}}

To validate the effectiveness of our AIGer approach, we conduct comparative experiments with the other SOTA approaches.
We selected several SOTA approaches that are currently recognized as the most advanced in AIG representation, with a focus on the DeepGate family—comprising DeepGate~\cite{li2022deepgate}, DeepGate2~\cite{shi2023deepgate2}, and DeepGate3~\cite{shi2024deepgate3}—known for their superior performance in modeling AIGs and other types of Boolean networks. In addition, we included PolarGate~\cite{liu2024polargate}, HOGA~\cite{deng2024less}, and FuncGNN~\cite{zhao2025funcgnn} in our comparison, which are open-source AIGNN frameworks that also demonstrate strong capabilities in AIG representation.
In summary, we use the aforementioned approaches as baselines to validate whether AIGer achieves optimal performance in AIG representation.

\subsection{\textbf{Training Details}}

During training, the experimental environment was deployed on a high-performance Linux server equipped with an NVIDIA A6000 GPU, which has 48GB of GPU memory.
To ensure fairness, training and testing were conducted on a single GPU in a consistent environment, with multiple experiments run to average results and mitigate the influence of different computational devices and environments. 
The Adam optimizer was used to accelerate learning, with an initial learning rate of 0.001, a fixed batch size of 256, and a default of 800 training epochs. 
Layer normalization was applied to the final layer to stabilize gradient flow and enhance model generalization, and an early stopping mechanism with a controllable patience value was used to prevent overfitting.
In this paper, to facilitate the reproduction of our work by researchers, the dataset and code used for training are publicly available~\cite{ourproject}.

\section{\textbf{Experimental Results}}\label{rs}

\subsection{\textbf{RQ1: Performance Comparison of SPP Task}}

\textbf{Approach. }
For the SPP task, which is widely used in circuit optimization and verification.
In the experiments, we selected AIGNN, which has demonstrated excellent performance, as the baseline model. 
We use MAE, MSE, and average training time per epoch T\_avg to assess the overall performance of the models.
In the experiment, for the DeepGate model, we trained and tested the complete model equipped with additional attention mechanisms and skip connections. For DeepGate2, we tested both single-stage training and two-stage training, ultimately selecting the two-stage training DeepGate2. For DeepGate3, we retained the multi-task pretraining strategy and sliding window mechanism. 
In the Table ~\ref{rq1} , L represents the number of layers in the PolarGate model, where we selected the 3-layer and 9-layer PolarGate models from the paper for experimentation.All training details were kept consistent with those in the paper.
For FuncGNN, we used the 3-layer feature aggregation model, which provided the best overall performance, for comparison.

\textbf{Results. }
Table~\ref{rq1} presents the training and testing results for the SPP task. From the data in the Table~\ref{rq1}, it can be concluded that the AIGer model significantly outperforms the existing baseline models across all three metrics. 
It demonstrates the best performance in terms of MAE, MSE, and T\_avg, indicating that AIGer currently achieves the most outstanding performance in the SPP task involving AIGs representation.

From the data in Table~\ref{rq1}, it can be concluded that AIGer demonstrates the best performance in the MAE and MSE metrics for the SPP task. In terms of the MAE metric, AIGer improves by 18.95\% compared to the best-performing FuncGNN, and up to 94.27\% compared to HOGA. In the MSE metric, AIGer shows an improvement of 44.44\% over the best FuncGNN and up to 97.67\% compared to HOGA.
Meanwhile, the AIGer model optimizes the training time, with its average training time reduced by 4.04\% compared to FuncGNN.
In fact, the AIGer model with 9 convolutional layers achieves even shorter training times and performs better than existing approaches. However, the AIGer model with 12 convolutional layers exhibits the best overall performance.

In summary, AIGer demonstrates the best performance in the SPP task, achieving improvements of 18.95\%, 44.44\%, and 4.04\% over the existing best models in the three metrics. 
This indicates that AIGer is currently the best-performing AIGNN for the SPP task, capable of the most accurate modeling and representation of AIGs data. 
In fact, AIGer's performance could further improve with deeper convolutional layers, but this would lead to significantly longer training times, which will be discussed in subsequent experiments.

\begin{table}[!t]
\centering  
\setlength{\tabcolsep}{8pt} 
\caption{The Statistics of AIG Datasets}
\label{dataset}  
\begin{tabular}{lccc}   
\toprule
\textbf{Benchmark} & \textbf{\#Subcircuits} & \textbf{\#Node} & \textbf{\#Level}  \\
\midrule
EPFL  & 828     & [52-341]  & [4-17]     \\
ITC99   & 7,560	& [36-1, 947] & [3-23]        \\
IWLS    & 1,281 & [41-2, 268] & [5-24]        \\
Opencores & 1,155	& [51-3, 214] & [4-18]      \\
\midrule
Total & 10,824 & [36-3, 214] & [3-24]     \\
\bottomrule
\end{tabular}
\end{table}

\begin{table}[!t]
\centering  
\setlength{\tabcolsep}{10pt} 
\caption{Performance of Various Models on the SPP Task}
\label{rq1}  
\begin{tabular}{lcccc}   
\toprule
\textbf{Model} & \textbf{MAE↓} & \textbf{MSE↓} & \textbf{T\_avg(s)↓}  \\
\midrule
DeepGate~\cite{li2022deepgate}	&	0.1105	& 0.0171	& 56.07	\\
DeepGate2~\cite{shi2023deepgate2}	&	0.0216	& 0.0032 & 40.12	\\
DeepGate3~\cite{shi2024deepgate3}	&	0.0245	& 0.0048 & 16.56	\\
HOGA~\cite{deng2024less}	&	0.1344	& 0.0215	& 5.82		\\
PolarGte(L=3)~\cite{liu2024polargate}	& 0.0435	& 0.0095	& 3.93		\\
PolarGte(L=9)~\cite{liu2024polargate}	& 0.0097	& 0.0011	& 6.33		\\
FuncGNN~\cite{zhao2025funcgnn}		& 0.0095	& 0.0009	& 3.71		\\
\textbf{AIGer}	& \textbf{0.0077}	& \textbf{0.0005}	& \textbf{3.56}		\\

\bottomrule
\end{tabular}
\end{table}

\subsection{\textbf{RQ2: Performance Comparison of TTDP Task}}

\textbf{Approach. }
The TTDP task is also a widely used evaluation metric in circuit optimization and verification. Similar to the SPP task, all models in this experiment employed the same training and testing strategies as those used for the SPP task. 
Notably, during the training of DeepGate2, we tested both single-stage and two-stage training. The two-stage training exhibited a clear performance improvement and delivered the best results, so we selected the two-stage trained DeepGate2 for this experiment.

\textbf{Results. }
Table~\ref{rq2} presents the results of different models under the TTDP task. From the data in the Table~\ref{rq2}, it can be concluded that the AIGer model generally outperforms existing baseline models, demonstrating the best performance in both MAE and MSE metrics. Additionally, the training time of AIGer is comparable to that of the optimal model.
The data in Table~\ref{rq2} shows that AIGer achieves the best performance in both MAE and MSE metrics for the TTDP task. AIGer outperforms the top-performing FuncGNN by 33.57\% in MAE and by up to 71.78\% compared to DeepGate3. In MSE, AIGer improves over FuncGNN by 14.79\% and by up to 72.6\% compared to DeepGate3. 
Although AIGer has a slightly longer computation time than FuncGNN, the difference is minimal, and its overall performance is superior. In fact, AIGer with 9 convolutional layers outperforms all baselines across all metrics, making it a good choice when computational resources are limited and time efficiency is prioritized.
In summary, AIGer demonstrates outstanding performance in the TTDP task, achieving improvements of 33.57\% and 14.79\% over the best existing models in the MAE and MSE metrics, respectively. 
This indicates that AIGer is the highest-performing AIGNN for the TTDP task. Considering its performance across both the SPP and TTDP tasks, AIGer is currently the best AIGNN, capable of providing the most accurate modeling and representation of AIGs data.

\subsection{\textbf{RQ3: The impact of different GNN propagation layers on the model}}

\textbf{Approach. }
The representation and modeling of AIGs heavily rely on the construction of GNNs, where feature processing and message propagation in the GNN network play a crucial role in model performance. 
To validate that the heterogeneous graph convolution layers in AIGer are the best GNN feature modeling and propagation layers, we replaced the heterogeneous graph convolution layer component in AIGer with other GNNs and conducted various experiments to verify that our constructed heterogeneous graph convolution network component is the optimal GNN propagation layer.

\begin{itemize}
    \item Graph Convolutional Networks (GCN)~\cite{kipf2016semi} are convolutional graph neural networks widely used for analyzing and processing graph data, and have previously been applied in AIGNN research.

    \item Relational Graph Convolutional Network (R-GCN)~\cite{schlichtkrull2018modeling} encodes relational features in multi-relational graph data and utilizes graph convolutional networks to process multi-relational data, enabling the understanding and learning of heterogeneous graphs with multiple relations.

    \item Graph Attention Networks (GAT)~\cite{velivckovic2017graph} introduce a self-attention mechanism to dynamically compute the attention weights between nodes and their neighbors, enabling weighted aggregation and representation learning of node features in graph data.

    \item Relational Graph Attention Network (R-GAT)~\cite{busbridge2019relational} introduces relation types to overcome the limitation of traditional GAT networks, which only consider node features during weighted aggregation of neighboring nodes. This enables R-GAT to handle multi-relational graphs and capture more complex information structures within the graph.

    \item Graph Sample and Aggregate (GraphSAGE)~\cite{hamilton2017inductive} is a spatial-based graph neural network model that randomly samples neighbors of target nodes and aggregates their feature information to learn from graph data.

    \item Variational Graph Auto-Encoder (VGAE)~\cite{kipf2016variational} maps graph data to a low-dimensional latent space using an encoder (e.g., GCN) to generate node embeddings, and reconstructs the adjacency matrix with a decoder, learning and representing the graph structure by minimizing reconstruction error.

    \item GENeralized Aggregation Networks (GEN)~\cite{li2020deepergcn} enhance information propagation in deep networks by incorporating generalized convolution operations.

\end{itemize}

\begin{table}[!t]
\centering  
\setlength{\tabcolsep}{10pt} 
\caption{Performance of Various Models on the TTDP Task}
\label{rq2}  
\begin{tabular}{lcccc}   
\toprule
\textbf{Model} & \textbf{MAE↓} & \textbf{MSE↓} & \textbf{T\_avg(s)↓}  \\
\midrule
DeepGate~\cite{li2022deepgate}		& 0.4305	& 0.2541	& 54.53	\\
DeepGate2~\cite{shi2023deepgate2}		& 0.3916	& 0.2445	& 47.86	\\
DeepGate3~\cite{shi2024deepgate3}		& 0.4348	& 0.2566	& 15.77	\\
HOGA~\cite{deng2024less}		& 0.3109	& 0.2267	& 5.78	\\
PolarGte(L=3)~\cite{liu2024polargate}		& 0.2521	& 0.2080	& 3.88	\\
PolarGte(L=9)~\cite{liu2024polargate}		& 0.2272	& 0.1038	& 5.77	\\
FuncGNN~\cite{zhao2025funcgnn}		& 0.1847	& 0.0825	& \textbf{3.61}	\\
\textbf{AIGer}	& \textbf{0.1227}	& \textbf{0.0703}	& 3.82	\\

\bottomrule
\end{tabular}
\end{table}

\begin{table}[!t]
\centering  
\setlength{\tabcolsep}{10pt} 
\caption{Model Performance under Different GNN Propagation Layer Numbers}
\label{rq3}  
\begin{tabular}{lcccc}   
\toprule
\textbf{GNN(L=12)} & \textbf{MAE↓} & \textbf{MAE↓} & \textbf{T\_avg(s)↓}  \\
\midrule
GCN~\cite{kipf2016semi}		& 0.0095	& 0.2002	& 4.53	\\
RGCN~\cite{schlichtkrull2018modeling}		& 0.0088	& 0.1452	& 5.02	\\
GAT~\cite{velivckovic2017graph}		& 0.0093	& 0.2036	& 5.21	\\
RGAT~\cite{busbridge2019relational}		& 0.0088	& 0.1688	& 6.12	\\
GraphSAGE~\cite{hamilton2017inductive}		& 0.0102	& 0.2243	& 4.03	\\
VGAE~\cite{kipf2016variational}		& 0.0146	 & 0.2437	& 4.11	\\
GEN~\cite{li2020deepergcn}		& 0.0114	& 0.2142	& 4.51	\\
\textbf{AIGerCov}	& \textbf{0.0077}	& \textbf{0.1227}	& \textbf{3.86}	\\

\bottomrule
\end{tabular}
\end{table}

In the experiment, apart from ensuring that all models run in the same environment, we defined the propagation depth of the GNN as 12 layers. 
In this experiment, we tested the MAE values of different GNN propagation depths on two tasks and the average training time on the SPP task to validate the superiority of the AIGer heterogeneous graph convolutional network component.

\textbf{Results. }
Table~\ref{rq3} shows the impact of different GNN propagation depths on the model. From the data in the table, it can be concluded that AIGer’s heterogeneous graph convolutional network exhibits the optimal performance in terms of both test results and training time consumption.

From the data in Table~\ref{rq3}, it can be concluded that in the SPP task, the original AIGer model outperforms the GNN with an RGCN replacement by 12.5\%, and achieves a 47.26\% improvement over the GAE. Additionally, the training time is reduced by up to 36.93\% compared to RGAT. In the TTDP task, AIGer shows a 15.5\% improvement over the GNN replaced by RGCN and a 49.65\% improvement over GAE. It is noteworthy that RGCN also performs well in the tasks; however, the default RGCN does not employ the base decomposition strategy to reduce the parameter count, nor does it utilize specific aggregation strategies for logical units. AIGer significantly improves on these aspects, leading to substantial performance gains in the model.

In summary, the AIGer heterogeneous graph convolutional network component demonstrates the best performance within the existing AIGNN message-passing mechanism, showing significant improvements compared to other traditional GNNs.

\subsection{\textbf{RQ4: The Impact of the Number of Convolutional Layers in AIGer on Training Accuracy and Training Efficiency}}

\textbf{Approach. }
Due to the deeply structured data inherent in AIGs, modeling AIGs is highly dependent on the depth of the model.
Generally, the deeper the AIGNN, the stronger its ability to retain information, enabling better adaptation to AIGs data. 
However, the feature propagation ability of AIGNN does not always improve with depth. Deep networks often face issues such as model over-compression, overfitting, and excessive parameter counts. 
On the other hand, shallow networks tend to lack stability and modeling capacity but offer faster training and inference speeds. In this experiment, we investigated the impact of different convolutional layer depths of AIGer on model performance.

For AIGer, training costs become significantly high when the number of layers exceeds 18, and computational resources cannot effectively support this. On the other hand, if the number of layers is fewer than 3, the data exhibits large fluctuations, leading to model instability. 
Therefore, we set the convolutional layer depths of the model to 3, 5, 9, 12, 15, and 18 for the experiment. 
During training, to validate the model's learning ability and generalization capability on small-scale training data in real-world scenarios, we tested four different data distributions for the training, validation, and test sets: 0.01-0.01-0.98, 0.02-0.02-0.96, 0.05-0.05-0.9, and 0.1-0.1-0.8. For example, split\_0.01 indicates a training-validation-test ratio of 0.01:0.01:0.98. 
By training and testing with different data proportions, we aimed to explore the model's robustness and generalization ability.
During testing, we selected the MAE metric for both the SPP and TTDP tasks and also recorded the average training time for the SPP task for comparison, facilitating observation of the experimental results.

\textbf{Results. }
Figure 4 shows the experimental results of AIGer with different numbers of convolutional layers, while Figure 5 illustrates the average training time for different convolutional layer depths. 
From the figures, it can be observed that as the number of convolutional layers increases from 3 to 18, the MAE metric steadily decreases across different training data proportions. 
When the number of convolutional layers exceeds 12, the variation becomes minimal, and the computational cost increases significantly, ultimately exhibiting exponential growth. Therefore, the model with 12 convolutional layers in AIGer provides the best overall performance. If computational resources are limited, the model with 9 layers also yields good results and saves training time.

\begin{figure}
    \centering
    \includegraphics[width=1\linewidth]{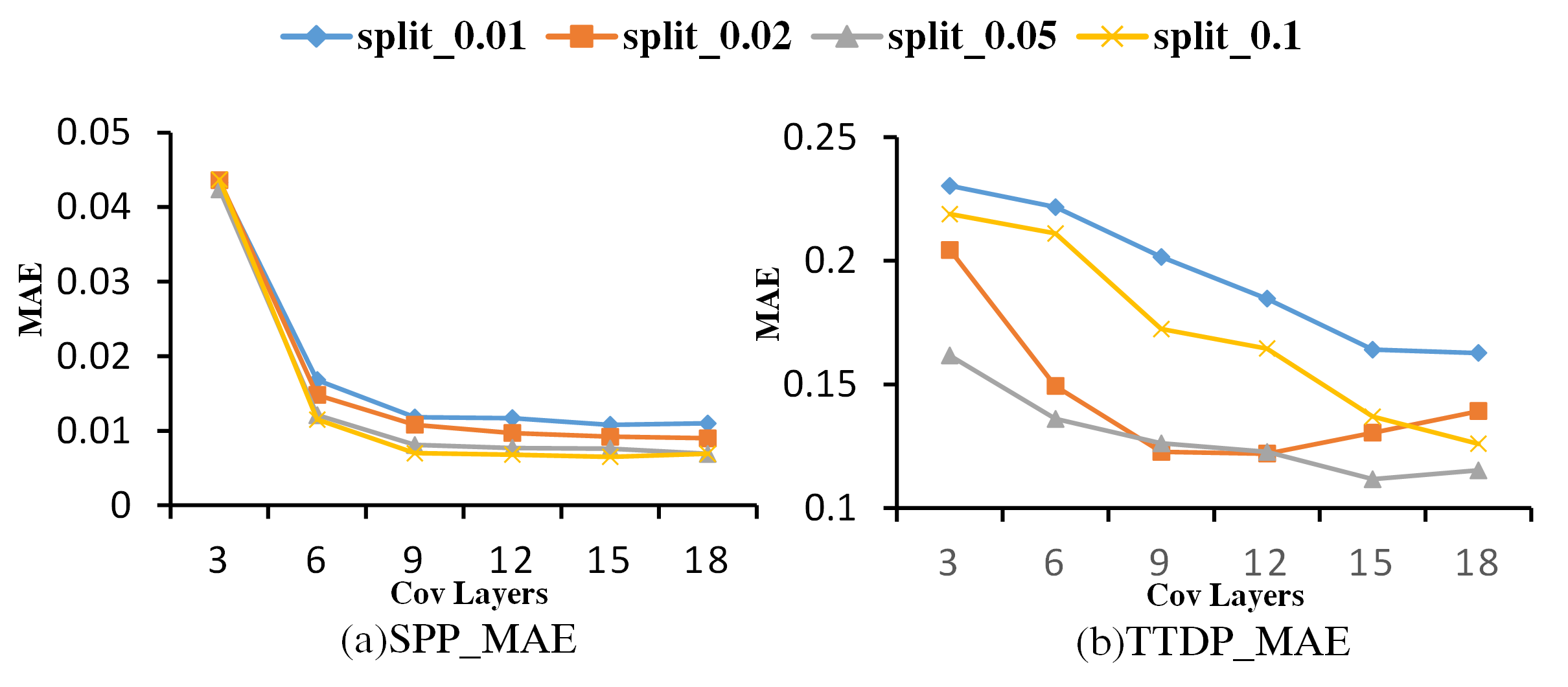}
    \caption{Experimental Results of AIGer with Different Numbers of Convolutional Layers}
    \label{line}
\end{figure}

\begin{figure}
    \centering
    \includegraphics[width=0.8\linewidth]{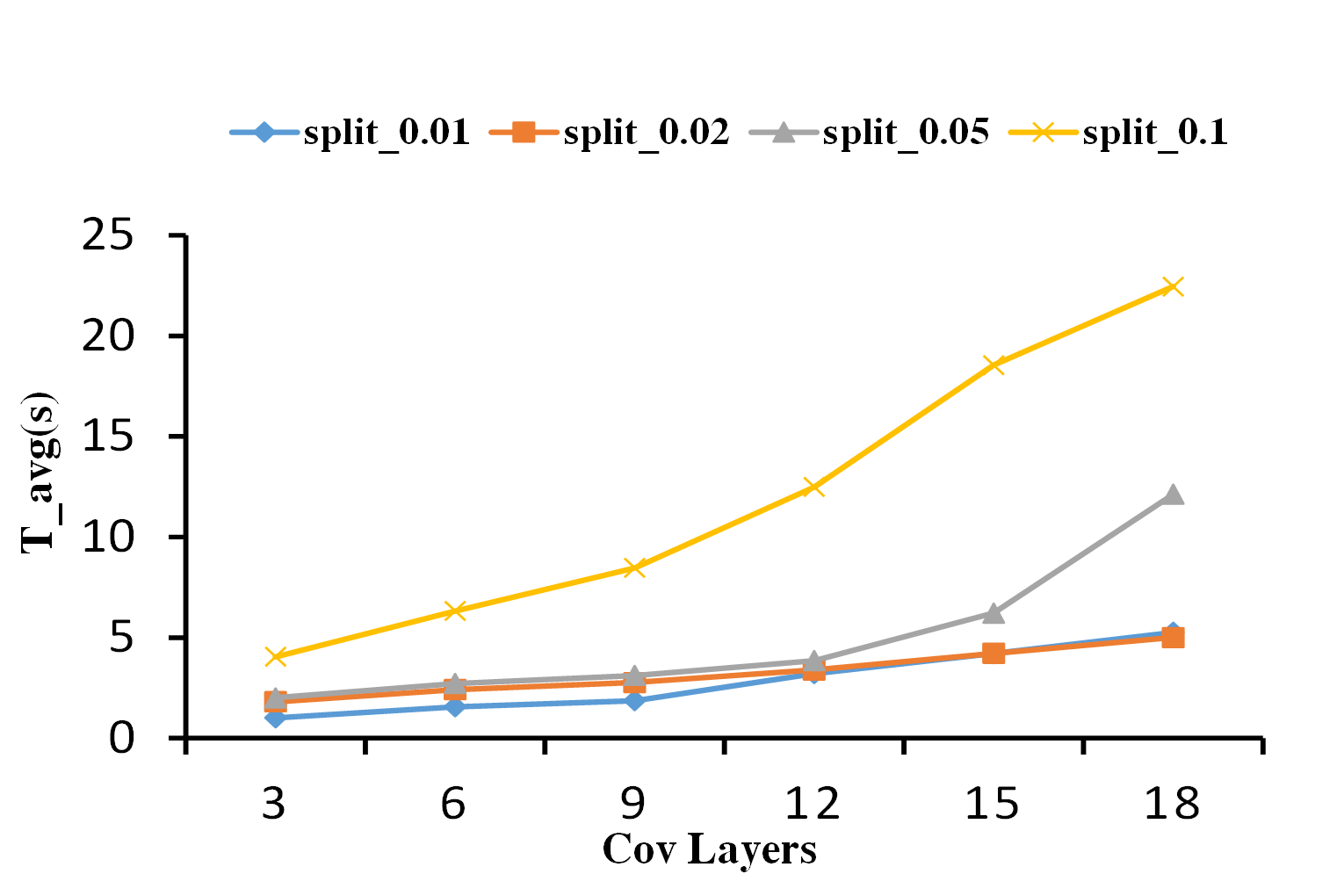}
    \caption{Average Training Time of AIGer with Different Numbers of Convolutional Layers}
    \label{line}
\end{figure}

\begin{table}[!t]
\centering  
\setlength{\tabcolsep}{10pt} 
\caption{The Impact of Each Component on the Model}
\label{rq5}  
\begin{tabular}{lcccc}   
\toprule
\textbf{Approach} & \textbf{SPP\_MAE↓} & \textbf{SPP\_MAE↓}   \\
\midrule
AIGer(Only\_Cov)		& 0.0285	& 0.2484	\\
AIGer(Only\_Emb) 		& 0.0197	& 0.2254	\\
AIGer(Single\_ Emb)		& 0.0127	& 0.2142	\\
AIGer(Sum\_Agg)		& 0.0088	& 0.1455	\\	
AIGer(NoDec)		& 0.0077	& 0.1228	\\
\textbf{AIGer}	& \textbf{0.0077}	& \textbf{0.1227}		\\

\bottomrule
\end{tabular}
\end{table}

\subsection{\textbf{RQ5: Ablation Study of the Components in AIGer}}

\textbf{Approach. }
Finally, we conducted an ablation study on the AIGer model to verify the necessity of each component and determine whether the current configuration is optimal. In this experiment, we tested the MAE metric for the SPP and TTDP tasks, using the default 12-layer GNN message-passing depth and validating models with different component combinations.

Table~\ref{rq5} represents models with different component configurations. AIGer(Only\_Cov) uses only the AIGer heterogeneous graph convolutional network component, directly feeding the features extracted by AIGer into the convolutional layers. AIGer(Only\_Emb) initializes multi-relation embeddings, using multi-layer MLP for feature propagation and aggregation. AIGer(Single\_Emb) performs initialization with unidirectional embeddings, without distinguishing the logical functions of nodes. AIGer(Sum\_Agg) employs the traditional weighted sum aggregation operation in the GNN message passing, without distinguishing node functions. AIGer(NoDec) does not use the base decomposition strategy to reduce parameter scale.
ok

\textbf{Results. }
From the results in Table~\ref{rq5}, the performance of AIGer’s components and their combined performance can be analyzed. 
When the initial node embedding component is removed, the overall performance on both tasks decreases by an average of 61.79\%. When the node initial embedding does not account for the logical features of nodes and only embedding operations are applied, the overall performance on both tasks decreases by 41.04\%. When the heterogeneous graph convolutional network component is removed and replaced with a traditional fully connected network, the performance on both tasks decreases by an average of 53.24\%. When the heterogeneous graph convolutional network component uses the traditional message-passing mechanism with weighted sum aggregation, the performance on both tasks decreases by 14.09\%. The base decomposition strategy is key to reducing parameter weights and accelerating learning without significantly affecting training results.

In conclusion, when the components of AIGer are used separately, they fail to effectively represent AIGs, showing inferior performance compared to AIGer. However, AIGer with its complete components achieves optimal performance and is currently the most performant AIGNN.

\section{\textbf{Conclusion}}

In this paper, we propose AIGer, a high-performance and lightweight model for modeling and representing AIGs data. 
AIGer consists of two key components: the node logical feature initialization embedding component and the AIGs feature learning network component.
The synergy between these two components enables the model to address the challenges faced by existing AIGNNs in simultaneously capturing both the Boolean logical semantics and hierarchical topological structure of AIGs, while also mitigating information loss and low training efficiency in long-distance propagation. 
The node logical feature initialization embedding component optimizes the initial representation of nodes in AIGs data, while providing an embedding space for different logical operators. 
In the AIGs feature learning network component, the designed convolutional network propagation mechanism and aggregation operators optimize the representation and transmission of feature information, while the base decomposition strategy reduces the model's parameters and training time.

AIGer outperforms existing baselines on both the SPP and TTDP tasks, while also optimizing the average training time.
The model demonstrates high practicality, making it applicable to various AIG tasks and extendable to a broader range of tasks in the EDA domain.



\bibliographystyle{IEEEtran}
\bibliography{GUIDANCE, IEEEabrv}
\vspace{12pt}

\end{document}